\documentclass{article}
\usepackage{spconf,amsmath,graphicx}

\usepackage{cite}

\usepackage[utf8]{inputenc} 
\usepackage[T1]{fontenc}    
\usepackage{hyperref}       
\usepackage{url}            
\usepackage{booktabs}       
\usepackage{amsfonts}       
\usepackage{nicefrac}       
\usepackage{microtype}      
\usepackage{xcolor}         
\usepackage{multirow}
\usepackage{pifont}
\usepackage{comment}
\usepackage{soul}
\title{Dialog act guided Contextual Adapter for \\ personalized speech recognition}
\name{\begin{tabular}{@{}c@{}}
Feng-Ju Chang,
Thejaswi Muniyappa,
Kanthashree Mysore Sathyendra, \\
Kai Wei, 
Grant P. Strimel, 
Ross McGowan
\end{tabular}}
\address{Alexa, Amazon, USA}
\begin{document}
\ninept
\maketitle
\begin{abstract}

Personalization in multi-turn dialogs has been a long standing challenge for end-to-end automatic speech recognition (E2E ASR) models. Recent work on contextual adapters has tackled rare word recognition using user catalogs. This adaptation, however, does not incorporate an important cue, the dialog act, which is available in a multi-turn dialog scenario. 
In this work, we propose a dialog act guided contextual adapter network. Specifically, it leverages dialog acts to select the most relevant user catalogs and creates queries based on both -- the audio as well as the semantic relationship between the carrier phrase and user catalogs to better guide the contextual biasing. On industrial voice assistant datasets, our model outperforms both the baselines - dialog act encoder-only model, and the contextual adaptation, leading to the most improvement over the no-context model: 58\% average relative word error rate reduction (WERR) in the multi-turn dialog scenario, in comparison to the prior-art contextual adapter, which has achieved 39\% WERR over the no-context model.

\end{abstract}

\section{Introduction}
\label{sec:intro}

Personalized speech recognition has gained considerable attention in recent years as industrial voice assistants (IVAs) rise in popularity.
To provide the best customer experience, 
an IVA should be able to recognize personalized requests correctly. For example, \textit{``drop in on [Proper Name]''}, where the name could be user defined words. 
 
End-to-end deep neural networks have become the mainstream approach for ASR in IVAs, due to their great capacity to learn the audio-to-text mapping without the alignment information between the audio and text transcript~\cite{chiu2018state}. These systems include the Connectionist Temporal Classification (CTC) \cite{Graves2006ConnectionistTC}, Listen-Attend-Spell (LAS)\cite{LAS2016}, Recurrent Neural Network Transducer (RNN-T) \cite{graves2012sequence}, and the transformer network \cite{Dong2018SpeechTransformerAN}. 
However, E2E ASR models face challenges in generalizing to recognize user-specific rare words,
such as contact names, proper nouns, and named entities due to the scarcity of paired audio-to-text training data~\cite{bruguier2016learning,sainath2018no,pundak2018deep,chen2019joint,bruguier2019phoebe}.
 
Prior works have explored various types of contextual information to improve personalized speech recognition~\cite{Mohri2002WeightedFT,williams2018contextual,liu2021domainaware,jain2020contextual,liu2020contextualizing,chang2021context,sathyendra2022contextual,pundak2018deep, gourav2021personalization}: The weighted finite state transducers (WFSTs)~\cite{Mohri2002WeightedFT} on the rare words \cite{williams2018contextual}; Domain-aware neural language models~\cite{liu2021domainaware}; Text metadata of video~\cite{jain2020contextual,liu2020contextualizing}, and the catalogs~\cite{chang2021context,sathyendra2022contextual} provided by speakers such as contact names and device names. 
The contexts are incorporated to the model either with the shallow fusion~\cite{gourav2021personalization,Zhao2019} or the neural biasing approaches \cite{bruguier2019phoebe,pundak2018deep,abro2019multi,chang2021context,sathyendra2022contextual}. The neural solutions have been shown outperforming the use of grammars or dynamic WFSTs in the shallow fusion.
In~\cite{chang2021context,sathyendra2022contextual}, the catalog entities are encoded by BiLSTM or BERT based encoders, and then integrated with the audio encoder via a multi-head attention based contextual biasing network. The model is taught to leverage user-specific contexts to improve the recognition of rare words. The performance of this biasing approach, however, may be hindered when irrelevant user catalogs or/and wrongly predicted carrier phrases bias the next predictions towards a wrong token.

\begin{figure}[t]
    \centering
    \includegraphics[width=0.95\linewidth]{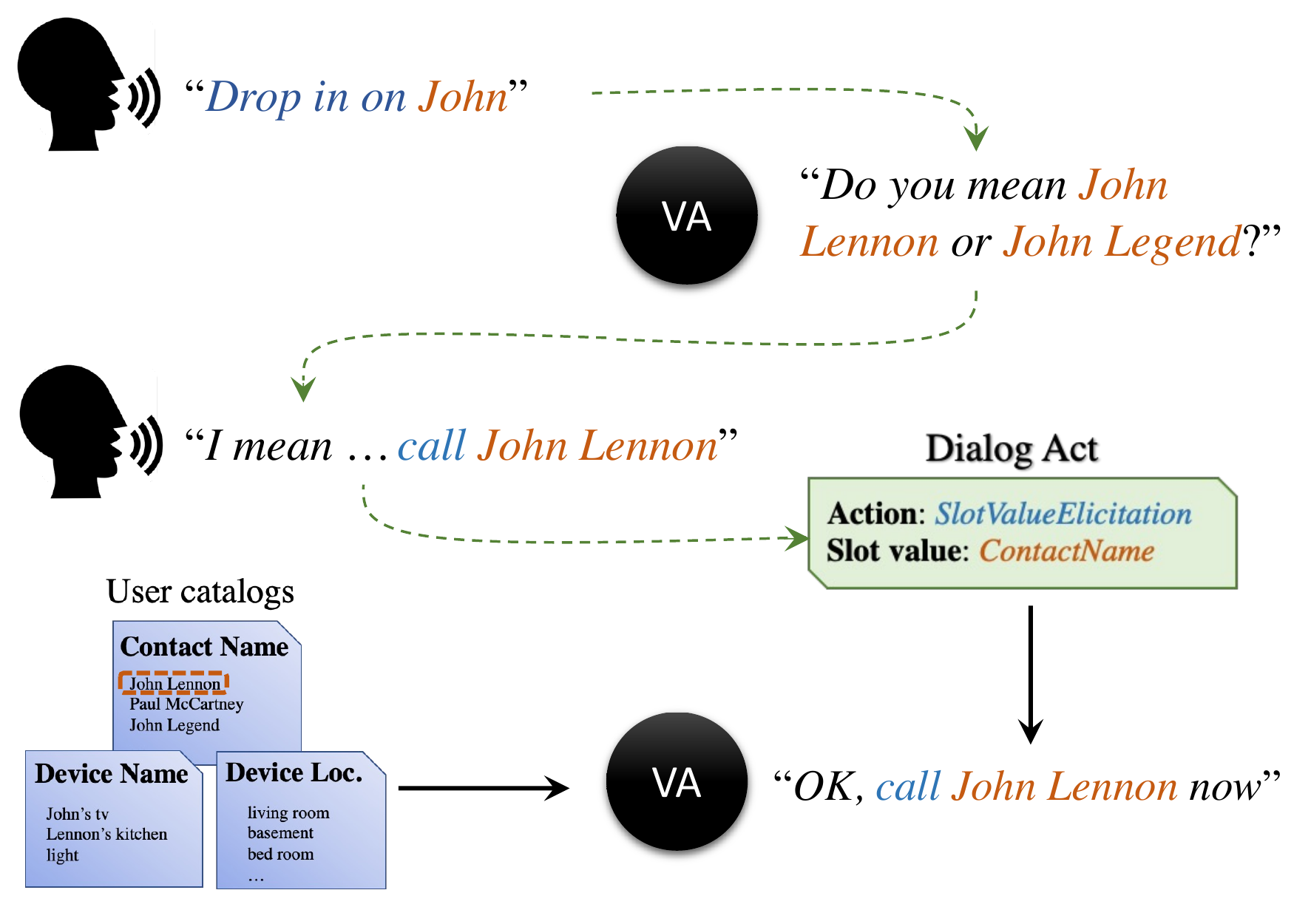}
    \vspace{-2.5mm}
    \caption{
    A high-level diagram illustrating our idea to leverage both dialog act and user catalogs to help a voice assistant (VA) recognizing proper nouns in a multi-turn dialog. Note that prior contextual adapters~\cite{chang2021context,sathyendra2022contextual} consider only user catalogs for contextual biasing.
    }
    \label{fig:da_eb_high_level}
    \vspace{-0.5cm}
\end{figure}

\begin{figure*}[t]
    \centering
    \includegraphics[width=0.7\linewidth]{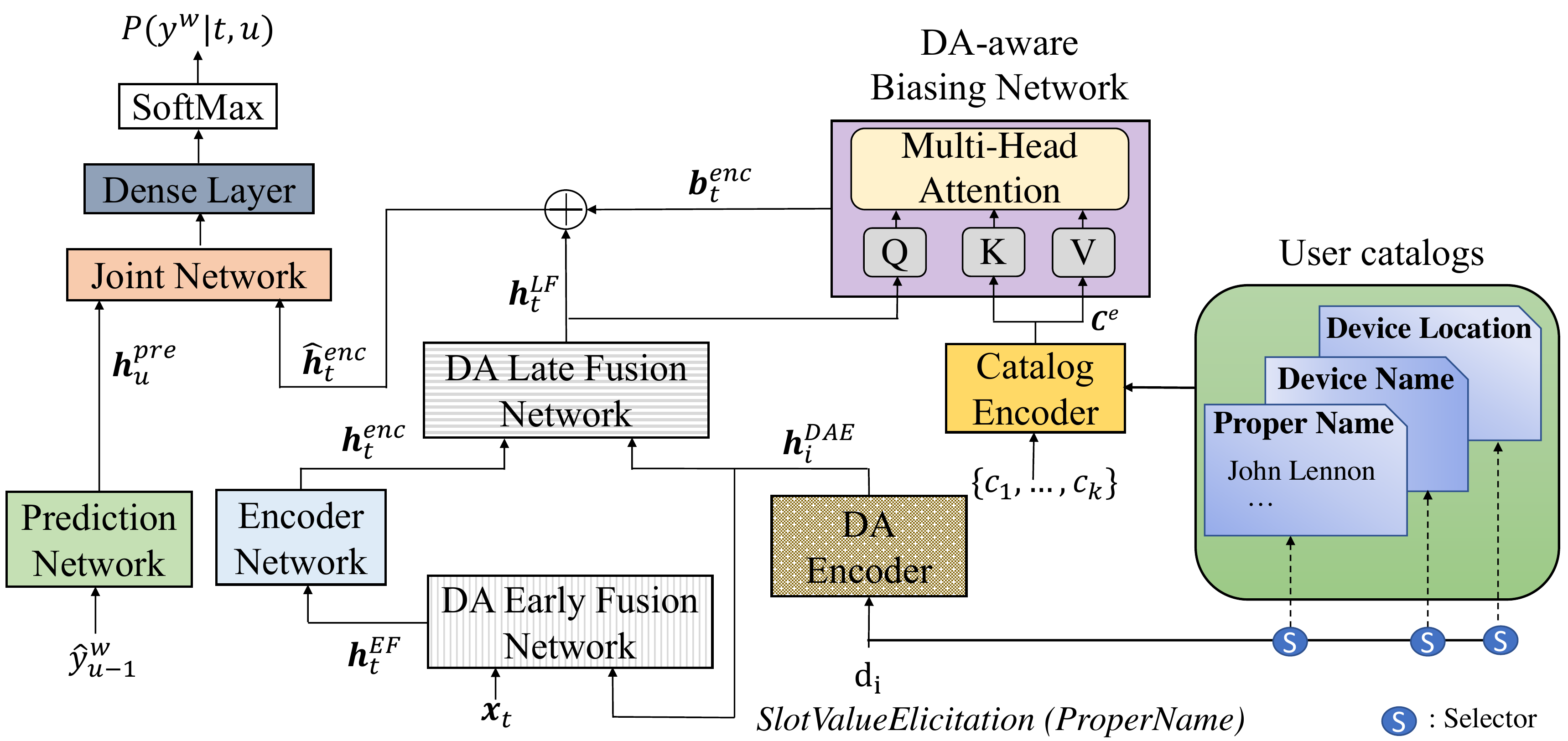}
    \vspace{-2.5mm}
    \caption{\textbf{The proposed DA guided contextual adapter.} The DAs are used for catalog selection and are fused with different levels of audio representations to build better queries for the biasing network.
    }
    \label{fig:ca_da}
    \vspace{-0.5cm}
\end{figure*}

To address the above issues, we propose to leverage both dialog act and user catalogs for the contextual biasing (Fig.~\ref{fig:da_eb_high_level}) and introduce a dialog act guided contextual adapter network (Fig.~\ref{fig:ca_da}). The dialog acts (DA) are used to identify the action and the slot value of spoken utterances in multi-turn dialogues (Fig.~\ref{fig:da_eb_high_level}). An example of DA is \textit{SlotValueElicitation(ProperName)}, which is associated with a dialog when a user tries to call someone. The DA can help the VA to elicit a contact name from the user catalog for the next turn, usually to clarify or confirm the entity. Given the semantic relationship between ``\textit{drop in on}'' or ``\textit{call}'' versus ``\textit{SlotValueElicitation}", DAs can help improve the prediction of the carrier phrases. We then use an attention based contextual adapter to 
bias the predictions towards the tokens in the user's \textit{[Contact Name]} catalog.
Note that prior work with DAs~\cite{wei2021attentive,kim2019gated,wu2020multistate,gupta2018efficient} primarily focus on the generic multi-turn utterance improvements, while the contextual biasing approaches~\cite{chang2021context,sathyendra2022contextual} were mainly dedicated to the single-turn named entity improvement. With the proposed network, we can bridge the gap between a single turn and multi-turn personalization in ASR. 


Our contributions are summarized as follows. First, we present the dialog act (DA) encoder (Section \ref{sec:da_enc}) and the fusion networks (Section \ref{sec:da_fusion_net}) to integrate the dialog act context with both the low level and high level acoustic representations, and improve the queries for the biasing network. Second, we use dialog acts to select the most relevant catalogs (Section \ref{sec:da_based_cs}), which directly influence the keys/values used in the biasing network. Third, we introduce a DA-aware biasing network, which takes the improved queries and keys/values above to bias toward the right catalog entities (Section \ref{sec:biasing_net}). Finally, we propose a two-stage adapter training and investigate the impacts of freezing and unfreezing DA encoder and fusion networks during the contextual biasing stage (Section \ref{sec:adapter_training}).

\section{Contextual Adapter}
\label{sec:bg}

\newcommand{\cnemb}{c^e_i}
\newcommand{\cn}{c_i}
\newcommand{\cni}[1]{c_{#1}}
\newcommand{\cniemb}[1]{c_{#1}^e}
\newcommand{\call}{C}
\newcommand{\cemball}{\boldsymbol{C}^e}
\newcommand{\attni}{\boldsymbol{\alpha}_t}
\newcommand{\attnit}{\alpha_i^t}
\newcommand{\dotprod}{\cdot}
\newcommand{\rootd}{\sqrt{d}}
\newcommand{\querylin}{\boldsymbol{W}^{q}}
\newcommand{\keylin}{\boldsymbol{W}^{k}}
\newcommand{\vallin}{\boldsymbol{W}^v}
\newcommand{\accousticquery}{q}
\newcommand{\semanticquery}{h_u^{pre}}
\newcommand{\encattncont}{b_t^{enc}}
\newcommand{\biasvector}{b}
\newcommand{\encbiasvector}{b^{enc}_t}
\newcommand{\decbiasvector}{b^{pre}_u}
\newcommand{\jointbiasvector}{b_{t,u}}
\newcommand{\jhat}{\hat{j}_{t,u}}
\newcommand{\maxcatsize}{K}
\newcommand{\htcap}{\hat{h}_t^{enc}}
\newcommand{\hucap}{\hat{h}_u^{pre}}
\newcommand{\nobias}{\textit{\textless no\_bias\textgreater} }
\newcommand{\encout}{h_t^{enc}}
\newcommand{\decout}{h_u^{pre}}
\newcommand{\decinp}{y_{u-1}}
\newcommand{\encinp}{x_{t}}
\newcommand{\encinpstart}{x_{0}}
\newcommand{\decinpstart}{y_{0}}
\newcommand{\catlen}{K}
\newcommand{\catembsize}{D}
\newcommand{\fullcatalog}{$\call = [\cni{1}, \cni{2}, \dots, \cni{K}]$}
\newcommand{\fullcatalogemb}{$\cemball = [\cniemb{1}, \cniemb{2} \dots \cniemb{K}]$}
\newcommand{\notype}{\textit{\textless no\_type\textgreater} }

The contextual adapter used in this work,~\cite{sathyendra2022contextual}, is built upon a type of streaming E2E ASR, RNN-T \cite{graves2012sequence}, which consists of an RNN based encoder, an RNN based prediction network, and a joint network. The encoder network produces high-level representations $\mathbf{h}_t^{enc}$ for the audio frames, while the prediction network encodes the previously predicted word-pieces and produces the output $\mathbf{h}_u^{pre}$. The joint network fuses $\mathbf{h}_t^{enc}$ and $\mathbf{h}_u^{pre}$ via the join operation followed by a series of dense layers with activations and a softmax function to obtain the probability distribution over word-pieces plus a $blank$ symbol. In~\cite{sathyendra2022contextual}, a catalog encoder is introduced to embed the user catalogs and types. In addition, a multi-head attention based biasing network is introduced to measure the relevance of the user catalog context using the encoder network outputs as queries. 
\section{Dialog Act Guided Contextual Adapter}
\label{sec:proposed}

In order to leverage DAs to guide the contextual adapter learning, we introduce four new components as shown in Fig.~\ref{fig:ca_da}: (1) DA Encoder, (2) DA Fusion Networks, (3) Catalog selection with DAs, and (4) DA aware Biasing Network, along with the two stage adapter training.

\subsection{DA Encoder}
\label{sec:da_enc}
\begin{figure}[t]
    \centering
    \includegraphics[width=1.0\linewidth]{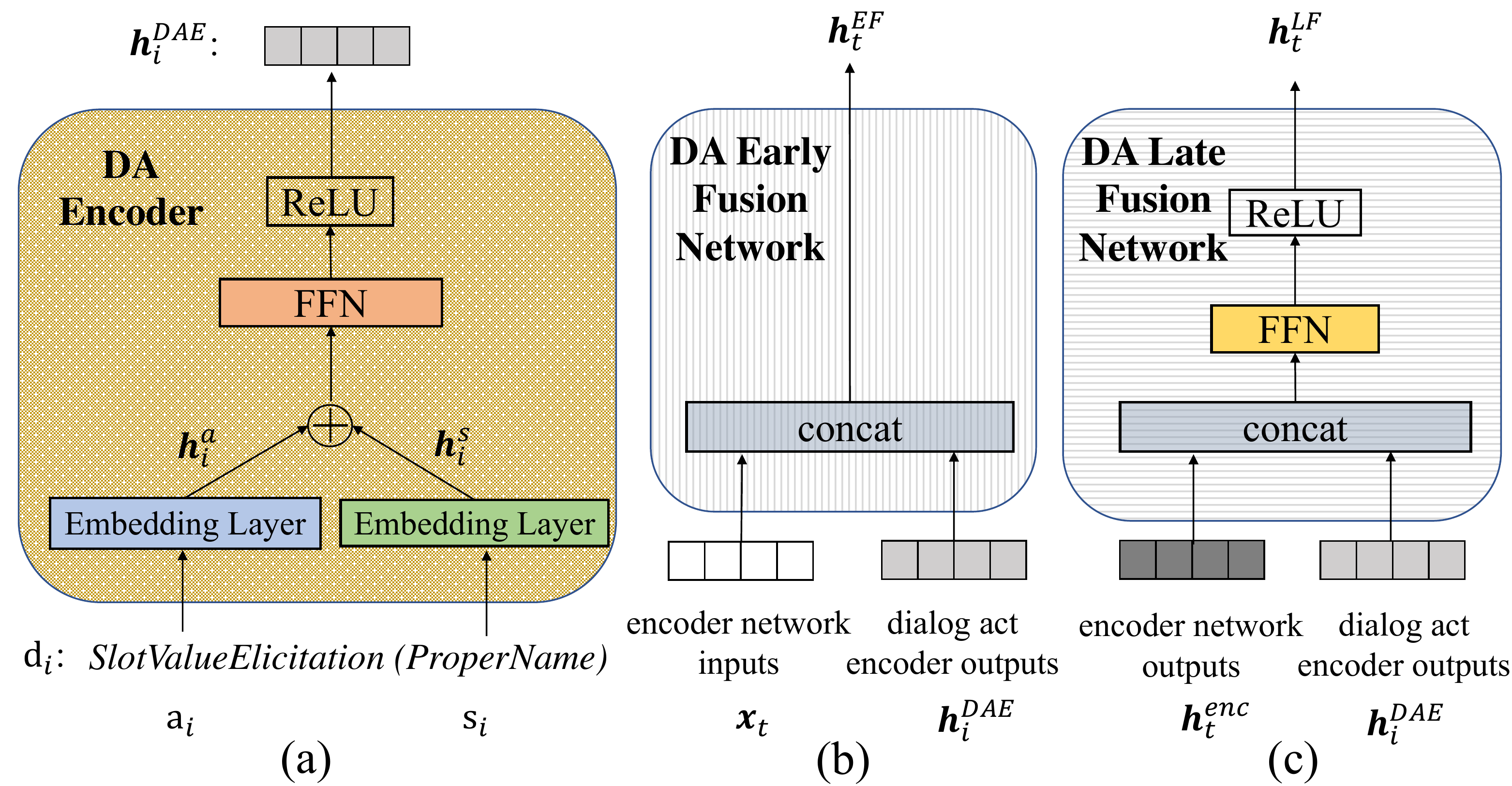}
    \vspace{-20pt}
    \caption{\textbf{The DA encoder and fusion networks}. (a) DA encoder, (b) DA early fusion network, and (c) DA late fusion network}
    \label{fig:da_enc_fusion}
    \vspace{-10.5mm}
\end{figure}

Each DA string, $\text{d}_i$, contains the action string $\text{a}_i$ and the slot string $\text{s}_i$, in the form of $\text{a}_i(\text{s}_i)$, e.g. \textit{SlotValueElicitation(ProperName)}, as shown in Figure~\ref{fig:da_enc_fusion}(a). Motivated by \cite{wei2021attentive}, we convert $\text{a}_i$ and $\text{s}_i$ by two embedding matrices into the action embeddings $\mathbf{h}_i^{a}$ and slot embeddings $\mathbf{h}_i^{s}$, with the same embedding size. Both embeddings are then fused via an element-wise addition followed by a Feed Forward Network (FFN) and the \text{ReLU} activation, denoted as $\sigma$, to produce the fixed dimensional DA embedding, $\mathbf{h}_i^{DAE} \in \mathbb{R}^{1 \times d_{da}}$: 

\begin{align}
    \vspace{-2.5mm}
    \mathbf{h}_i^{DAE} = \sigma(\text{FFN}(\mathbf{h}_i^{a}+\mathbf{h}_i^{s}))
    \label{eq:da_enc}
\end{align}

\subsection{Fusion of DAs and Multi-level Audio Embeddings}
\label{sec:da_fusion_net}

After we obtain the DA embedding, $\mathbf{h}_i^{DAE}$, it is combined with two levels of the audio representations using one of the three approaches presented below:

\textbf{Early Fusion}: We simply concatenate $\mathbf{h}_i^{DAE}$ with the input audio features per frame, $\mathbf{x}_t$, see Figure~\ref{fig:da_enc_fusion}(b).
\begin{align}
    \mathbf{h}_t^{EF} = \text{Concat}(\mathbf{x}_t, \mathbf{h}_i^{DAE})
    \label{eq:early_fusion}
\end{align}

\textbf{Late Fusion}: The encoder network output, $\mathbf{h}_t^{enc}$ is first concatenated with $\mathbf{h}_i^{DAE}$. The concatenated output is then transformed by a FFN with the \text{ReLU} activation, denoted as $\sigma$. See Figure~\ref{fig:da_enc_fusion}(c).
\begin{align}
    \mathbf{h}_t^{LF} = \sigma(\text{FFN}(\text{Concat}(\mathbf{h}_t^{enc}, \mathbf{h}_i^{DAE})))
    \label{eq:late_fusion}
\end{align}

\textbf{Early-Late Fusion}: In this approach, we inject $\mathbf{h}_i^{DAE}$ jointly with the input audio (Equation~\ref{eq:early_fusion}) and with the encoder output (Equation~\ref{eq:late_fusion}) as shown in Figure~\ref{fig:ca_da}. Note that the DA encoder is shared in this approach.

\subsection{Catalog Selection with DAs}
\label{sec:da_based_cs}

In addition to fusing the DA embeddings with the audio representations (Section~\ref{sec:da_fusion_net}), we also investigate the selection of catalogs based on the DA string $\text{d}_i=\text{a}_i(\text{s}_i)$, during training. Specifically, the slot string $\text{s}_i$ in $\text{d}_i$ is used for selecting the catalog for biasing. We consider three types of catalogs in this work: contact names, device names, and device locations. By matching $\text{s}_i$ with the catalog type, we can summarize the selection rule as follows:
\vspace{-0.5mm}
\begin{align}
    \small
     \{c_1,...c_K\} \in \begin{cases}
			\text{proper names catalog}, & \text{if $\text{s}_i$ = \textit{ProperName}}\\
			\text{device names catalog}, & \text{if $\text{s}_i$ = \textit{DeviceName}}\\
			\text{device locations catalog}, & \text{if $\text{s}_i$ = \textit{DeviceLocation}}\\
            \text{all three catalogs}, & \text{otherwise}
		 \end{cases} \nonumber
\end{align}

Using this approach, we can prevent irrelevant catalogs from deteriorating biasing network training.

\subsection{DA Aware Biasing Network}
\label{sec:biasing_net}
The biasing network learns which catalog entities to bias toward. Inspired by~\cite{chang2021context,sathyendra2022contextual}, we employ cross-attention to attend over the catalog embedding matrix $\cemball \in \mathbb{R}^{K \times D_c}$ based on the input query $\mathbf{q}$, which comes from the DA late fusion network, $\mathbf{h}_t^{LF} \in \mathbb{R}^{1 \times D_a}$ (Equation~\ref{eq:late_fusion}). Here $K$, $D_c$, $D_a$ are the total number of catalog entities, the catalog embedding size, and the audio embedding size respectively. This query intuitively encodes both the audio and the DA information -- the semantic relationship between the carrier phrase and the user catalogs.

The query and the catalog entity embeddings are first projected via $\querylin \in \mathbb{R}^{D_a \times d}$ and $\keylin \in \mathbb{R}^{D_c \times d}$ to dimension $d$. An attention score vector $\attni$ for catalog entity embeddings is computed by the scaled dot product attention mechanism~\cite{NIPS2017_3f5ee243} as follows: 
\begin{align}
    \attni = \text{Softmax} \left(\frac{(\cemball \keylin) (\mathbf{h}_t^{LF}\querylin)^T}{\rootd} \right)
\end{align}
where $\attni \in \mathbb{R}^{K \times 1}$. The biasing vector $\mathbf{b}_t^{enc}$ is then computed as $\mathbf{b}_t^{enc} = (\attni)^T (\cemball \vallin) \in \mathbb{R}^{1 \times D_a}$, where ${\boldsymbol{W}^{v}} \in \mathbb{R}^{D_c \times D_a}$. The biasing vector $\mathbf{b}_t^{enc}$ is then combined with the $\mathbf{h}_t^{LF}$: $\hat{\mathbf{h}}_t^{enc} = \mathbf{h}_t^{LF} \oplus \mathbf{b}_t^{enc}$.

While the audio representation enables the biasing network to bias towards entities based on acoustic similarity, the DA helps the biasing network attend over the right \textit{type} of entity. For instance, if the DA is \textit{SlotValueElicitation(DeviceName)}, the biasing network can be guided to bias towards entities that are of type \textit{DeviceName}.

\subsection{Two-Stage Training: DA-Aware RNN-T Training followed by Contextual Adaptation}
\label{sec:adapter_training}

While improving user-specific word recognition, a desirable E2E ASR should be able to maintain the performance on general speech recognition. To this end, we adopt the two-stage training similar to \cite{sathyendra2022contextual}. We first pre-train the DA-Aware RNN-T, i.e. RNN-T (the encoder network, prediction network, joint network, dense layer) augmented with the DA encoder and DA fusion networks on the live IVA distribution of utterances. In the second stage of contextual adaptation on user catalogs, we keep the RNN-T weights frozen while training the catalog encoder and the biasing network on upsampled personalized data mixed with the original general data under two conditions: (1) Freeze the DA encoder (DA Enc.) and fusion network (FN), (2) Unfreeze/ keep finetuning the DA Enc. and FN.
\section{Experiments}

\subsection{Datasets and Evaluation Metrics}
\label{sec:dataset}

We use in-house IVA datasets where audio to transcription paired utterances are de-identified and randomly sampled from more than 20 domains such as \textit{Global}, \textit{Communications}, \textit{SmartHome}, \textit{Weather}, and \textit{Music}. We consider the top 49 frequently occurring DAs in the IVA data sets including the \textit{DefaultDialogAct} (usually associated with the first-turn utterances) and the non-default DAs (associated with more follow-up turns)\footnote{There does not exist an equivalent, publicly available contextual dataset containing both user catalogs and dialog acts in multi-turn dialogue scenario.}. 114k hours of data are used to pre-train the DA-Aware RNN-T. For training the catalog encoder and the biasing network, we use approximately 290 hours of data, containing a mix of \textit{user-specific} and \textit{general} training data with the ratio of 1.5:1 as suggested in~\cite{sathyendra2022contextual}. \textit{User-specific} datasets contain utterances with contact names, device names, and/or device locations. \textit{General} utterances are sampled from the original training data distribution. We evaluate the models and report results on multiple test sets including a 20 hour \textit{User-specific} dataset and a 75 hour \textit{General} dataset.
The \textit{User-specific} test set is further split into the default DA and non-default DA set for further comparisons.
To evaluate our model on individual turns of a dialog, we further created turn-wise test sets, which contain \textit{user-specific} and non-default DA utterances from the first turn (turn 1), the second turn (turn 2) and the third turn (turn 3). 

The relative word error rate reduction (WERR) is used throughout the experiments to summarize the overall, slot-type-wise, or turn-wise ASR performances. Given a model A's WER ($\text{WER}_A$) and a baseline B's WER ($\text{WER}_B$), the WERR of A over B can be computed by $(\text{WER}_B - \text{WER}_A) / \text{WER}_B$; a higher WERR indicates a better WER. We use three types of catalogs (Section~\ref{sec:da_based_cs}). The maximum number of catalog entities fed into the catalog encoder is set to $K=500$, and contains the correct \textit{user-specific} entities. 

\subsection{Experimental Setup}
\subsubsection{Model Configurations}
The DA guided contextual adapter has the following configuration. The encoder network has 5 LSTM layers and the prediction network is composed of 2 LSTM layers, both with 736 units followed by a feed-forward network of 512 units. The joint network is a fully-connected feed-forward component with one hidden layer followed by a $\tanh$ activation function. The DA encoder contains a 49-dim embedding layer followed by a feed-forward network of 64 units (i.e., $d_{da}=64$). The DA late fusion network has a feed-forward network of 512 units. The catalog encoder is a BiLSTM layer with 128 units (each for forward and backward LSTMs) with input size 64. The final output from the catalog encoder is projected to 64-dim (i.e., $D_{c}=64$). The biasing network projects the query, key and values to 64-dimensions (i.e., $d=64$). The resulting biasing vector is then projected to the same size as the encoder output, 512-dim (i.e., $D_{a}=512$).

We compare our model to the following baselines: (1) \textbf{No Context}: This is a vanilla RNN-T\cite{graves2012sequence}. (2) \textbf{DA only}: This is an RNN-T using only DAs integrated with the input audio features~\cite{wei2021attentive}, and (3) \textbf{CA}: This is an RNN-T based Contextual Adapter (CA) using only user catalogs ~\cite{sathyendra2022contextual}. All the models including the proposed one contain $\sim 35$ million parameters.

\begin{table}[t]
\centering 
\small
\tabcolsep=0.1cm
\caption{WERRs (\%) over No-context baseline~\cite{graves2012sequence}: Comparisons for DA guided CA, DA only, and CA models on IVA \textit{User-specific} and \textit{General} test sets. A higher number indicates a better WER.}
\label{tab:wers_on_p2a}
\resizebox{1.0\linewidth}{!}{%
\begin{tabular}{c|c|c|c|c}
\toprule
Model & DA Fusion & DA Enc.\&FN & User-Specific & General \\ \midrule
No context~\cite{graves2012sequence} & N/A & N/A & baseline & baseline \\ \hline
DA Only~\cite{wei2021attentive} & N/A & N/A & 1.49 & -0.95 \\ \hline
CA~\cite{sathyendra2022contextual} & N/A & N/A & 28.23 & -0.59 \\ \hline
\multirow{6}{*}{\textbf{DA guided CA (Ours)}} & Early Fusion  & freeze & 29.72 & -1.02 \\
 &   & un-freeze & 28.83 & -0.66 \\ \cline{2-5}
 & Late Fusion  & freeze & 28.75 & -1.31 \\
 &   & un-freeze & 30.61 & -2.58 \\ \cline{2-5}
 & Early-Late Fusion & freeze & 30.53 & -2.3 \\
 &   & un-freeze & \textbf{32.39} & -2.5 \\
 \bottomrule
\end{tabular}}%
\vspace{-4.5mm}
\end{table}

\subsubsection{Input, Output, and Learning Rate Configurations}
The input audio features consists of 64-dim LFBE features, which are extracted every 10 ms with a window size of 25 ms from audio samples. The features of each frame are then stacked with the left two frames, followed by a downsampling of factor 3 to achieve low frame rate, resulting in 192 feature dimensions.
The subword tokenizer \cite{sennrich-etal-2016-neural, kudo2018subword} is used to create tokens from the transcriptions; we use 4000 tokens in total. 
We trained the RNN-T, the DA encoder, and the DA fusion network by minimizing the RNN-T loss~\cite{graves2012sequence} using the Adam optimizer \cite{kingma2014adam}, and varied the learning rate following \cite{Dong2018SpeechTransformerAN,NIPS2017_3f5ee243}.with the starting LR = 1.5e-7 and ramping up for $3000$ steps to reach LR = 4e-4. The LR is then held constant for $150K$ steps after which there is an exponential decay. For training the catalog encoder and biasing network in the CA stage, we use the Adam optimizer with LR = 1e-3 trained to convergence with early stopping.

\subsection{Results}

\begin{table}[t]
\centering
\small
\tabcolsep=0.1cm
\caption{WERRs (\%) over No-context baseline~\cite{graves2012sequence} on Non-Default/Default DA splits of the \textit{User-specific} test set. A higher number indicates a better WER.}
\label{tab:wers_on_da}
\resizebox{1.0\linewidth}{!}{%
\begin{tabular}{c|c|c|c|c}
\toprule
Model & DA Fusion & DA Enc.\&FN & Non-Default DA & Default DA \\ \midrule
No context~\cite{graves2012sequence} & N/A & N/A & baseline & baseline \\ \hline
DA Only~\cite{wei2021attentive} & N/A & N/A & 13.81 & -1.27 \\ \hline
CA~\cite{sathyendra2022contextual} & N/A  & N/A & 29.91 & 27.85 \\ \hline
\multirow{6}{*}{\textbf{DA guided CA (Ours)}} & Early Fusion  & freeze & 40.35 & 27.77 \\
 &    & un-freeze & 39.71 & 26.84 \\ \cline{2-5}
 & Late Fusion  & freeze & 39.01 & 27.00 \\
 &    & un-freeze & 39.54 & 29.03 \\ \cline{2-5}
 & Early-Late Fusion & freeze & 40.04 & 28.69 \\
 &  & un-freeze & \textbf{41.56} & \textbf{30.46} \\
 \bottomrule
\end{tabular}}%
\vspace{-5.5mm}
\end{table}

\begin{figure}
    \centering
    \includegraphics[width=1.0\linewidth]{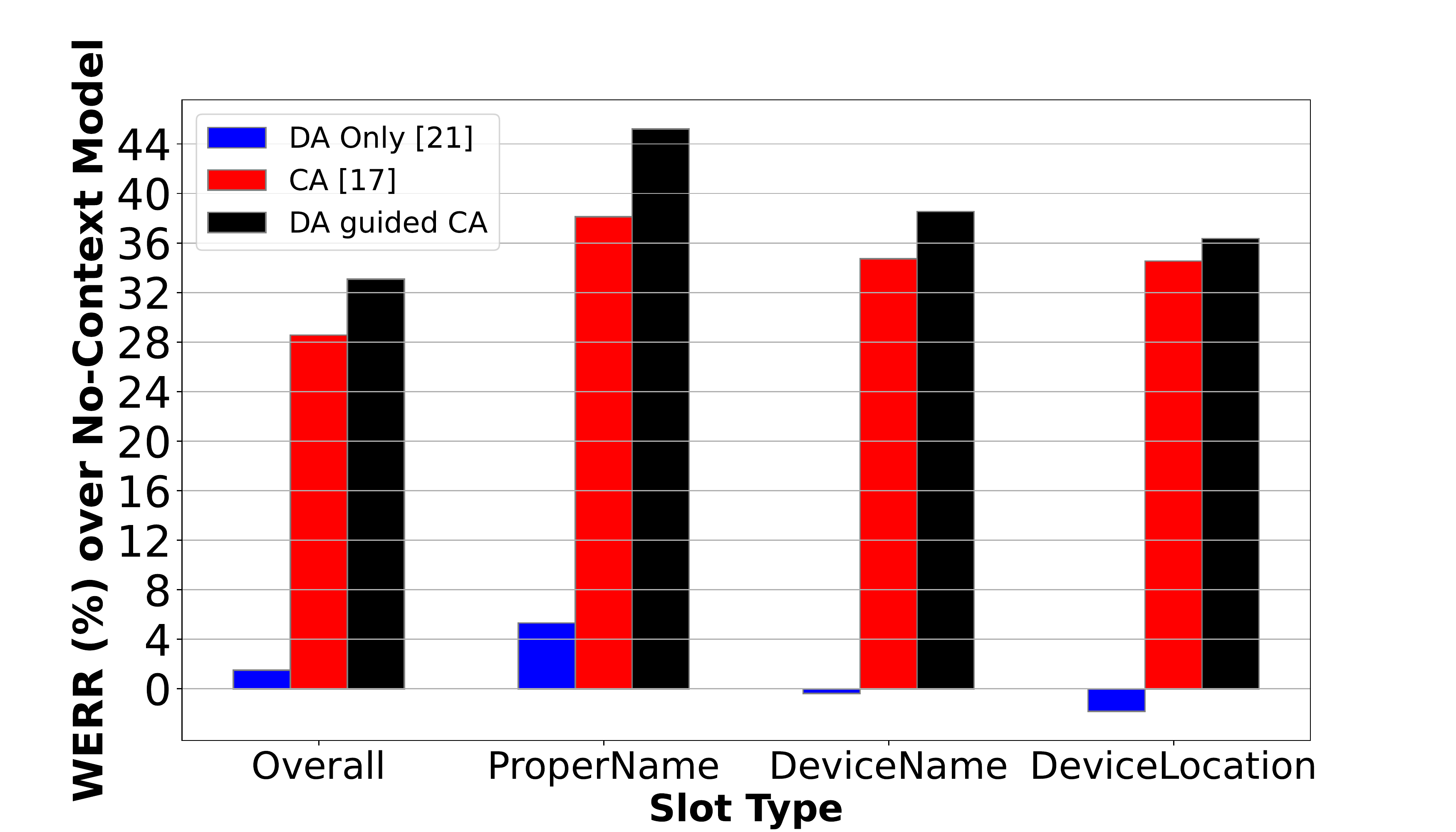}
    \vspace{-14pt}
    \caption{Slot WERRs (\%) over No-Context model~\cite{graves2012sequence} on the IVA \textit{User-specific} test set.}
    \label{fig:slot_werrs}
    \vspace{-10pt}
\end{figure}


Table~\ref{tab:wers_on_p2a} presents the WERRs of the DA guided CA and baseline approaches over the no-context model for the \textit{User-specific} and \textit{General} test sets. The DA guided CA 
outperforms the no-context model by $32.39\%$ relative improvement, while the DA only model and CA achieve $1.49\%$, and $28.90\%$ WERRs respectively on the \textit{User-specific} set. 
While unfreezing the DA Enc. and FN boosts the performance on the \textit{User-specific} sets, it is more prone to overbiasing, resulting in more degradations on the \textit{General} set compared to freezing their weights: -1.31\% WERR (freeze) vs. -2.58\% WERR (unfreeze) for the late dialog act fusion, and -2.3\% WERR (freeze) vs. -2.5\% WERR (unfreeze) for the early-late dialog act fusion in Table~\ref{tab:wers_on_p2a}.
We also observe that bypassing the catalog selection with DAs on the best performing DA guided CA, the WERR over No-Context model drops from $32.39\%$ to $31.43\%$, which indicates selecting catalogs and filtering catalog embeddings, i.e. keys/values, to the most relevant ones, leads to better biasing.

In Table~\ref{tab:wers_on_da}, we show the WERRs on non-default DA and default DA splits. The DA guided CA, again, improves WER over the No-Context model the most ($41.56\%$), compared to the WERRs achieved by the DA only ($13.81\%$) and CA ($29.91\%$) on the non-default DA split. The huge improvements seen in the non-default DA split can be explained by the fact that the \textit{user-specific} utterances are usually associated with non-default DA. 

We further compute the slot WERRs on three major \textit{user-specific} slots in Fig.~\ref{fig:slot_werrs}
along with the overall performances across all 131 slots. The DA guided CA model shows the most relative improvements in terms of Overall Slot WER against the No-Context model ($33.08\%$), compared to the WERRs of DA Only ($1.5\%$), and CA ($28.57\%$) respectively. Here, we can see the improvements on the contact name slot are the highest across all models. The proposed model achieves the most improvement ($45.21\%$ relative). Notably, although DA Only model performs slightly worse than the No-Context model for the device name and device location slots ($-0.38\%$ and $-1.82\%$ WERRs), DA guided CA still leads to better relative improvement over No-Context model, compared to the CA's improvement for the device name ($38.55\%$ vs. $34.73\%$) and device location slots ($36.36\%$ vs. $34.55\%$) respectively.

\begin{figure}
    \centering
    \includegraphics[width=0.9\linewidth]{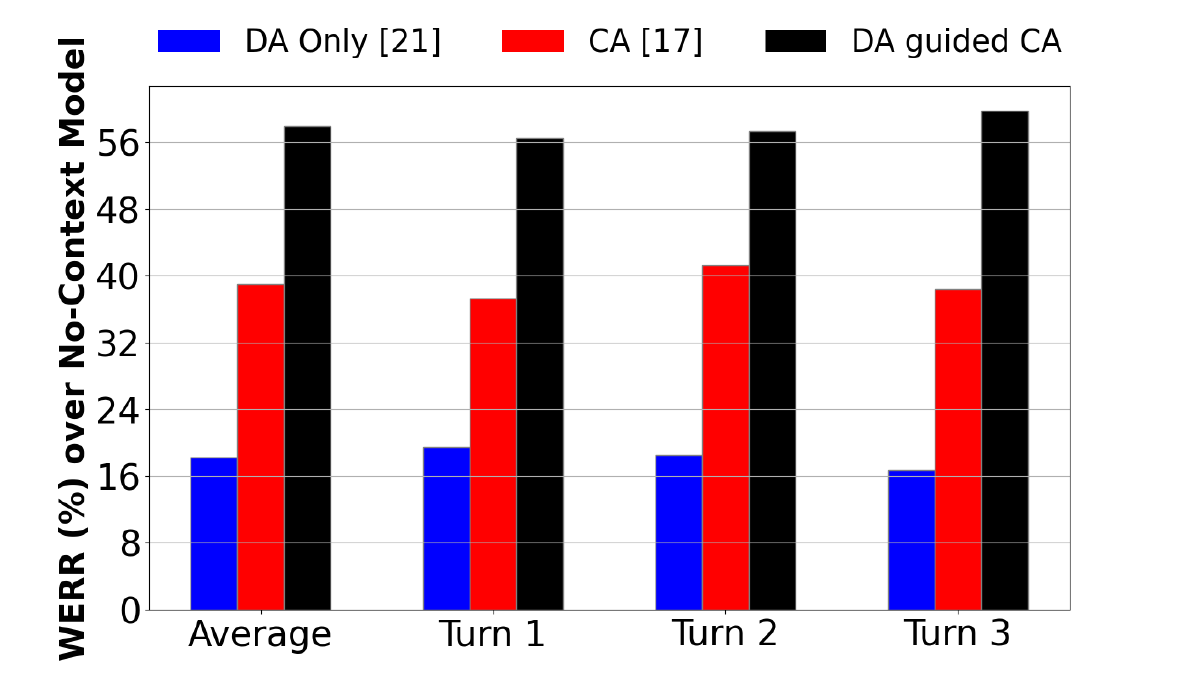}
    \vspace{-14pt}
    \caption{Turn-wise WERRs (\%) over No-context baseline~\cite{graves2012sequence} on the IVA \textit{User-specific} multi-turn test set.}
    \label{fig:turn_wise_werrs}
    \vspace{-10pt}
\end{figure}

Finally, we report the results on the turn-wise
test sets in Fig.~\ref{fig:turn_wise_werrs}.
The average of WERRs across turns are presented as well. While the DA Only model and CA have significant improvements, 
DA guided CA leads to the best improvements on average ($58\%$), turn 1 ($56.5\%$), turn 2 ($57.3\%$), and turn 3 ($59.8\%$) over the No-Context model. This shows that the DA guided CA model significantly improves performances for multi-turn personalized ASR over the baselines.
\section{Conclusion}
We proposed a dialog act guided contextual adapter approach to address personalized ASR in multi-turn dialogs. We leverage DAs to short list the most relevant catalogs and create better queries to guide the biasing network. The experimental results on IVA test sets show that the DA guided CA achieves on average $58\%$ WER relative improvements over the no-context model on the \textit{user-specific} multi-turn test set, in comparison to the prior-art contextual adapter model which achieved $39\%$ over the no-context model.


\label{sec:refs}

\bibliographystyle{IEEEbib}
\small \bibliography{refs}

\end{document}